\documentclass[preprint,12pt]{elsarticle}

\usepackage{color}
\usepackage{url}
\usepackage[colorlinks=true,linkcolor=blue,urlcolor=blue,citecolor=blue]{hyperref}
\usepackage{tabularx}

\usepackage{amssymb}
\usepackage{amsmath}
\journal{arXiv}

\begin{document}

\begin{frontmatter}



\title{Advancing Vehicle Plate Recognition: Multitasking Visual Language Models with VehiclePaliGemma }

\author[label1]{Nouar AlDahoul}
% \ead{naa9497@nyu.edu}
\author[label2]{Myles Joshua Toledo Tan}
\author[label3]{Raghava Reddy Tera}
\author[label4]{Hezerul Abdul Karim}
\author[label5]{Chee How Lim}
\author[label5]{Manish Kumar Mishra}
\author[label1]{\\Yasir Zaki\texorpdfstring{\corref{cor1}}{}}
% \ead{yasir.zaki@nyu.edu}

% \author{Nouar AlDahoul, Myles Joshua Toledo Tan, Raghava Reddy Tera, Hezerul Abdul Karim, Chee How Lim, Manish Kumar Mishra, Yasir Zaki*} %% Author name

\affiliation[label1]{organization=Computer Science Department, New York University Abu Dhabi,city=Abu Dhabi,country=UAE}    

\cortext[cor1]{Corresponding author.
E-mail address: yasir.zaki@nyu.edu}

\affiliation[label2]{organization=Department of Electrical and Computer Engineering, Herbert Wertheim College of Engineering, University of Florida,
            city=Florida,
            country=USA}

\affiliation[label3]{organization=Yo-Vivo Corporation, city= Bacolod City, Negros Occidental, country=Philippines}

\affiliation[label4]{organization=Faculty of Engineering, Multimedia University, city=Cyberjaya, Selangor,country=Malaysia}

\affiliation[label5]{organization=Tapway Sdn Bhd, Petaling Jaya, city=Selangor, country=Malaysia}

%% Abstract
\begin{abstract}
License plate recognition (LPR) involves automated systems that utilize cameras and computer vision to read vehicle license plates. Such plates collected through LPR can then be compared against databases to identify stolen vehicles, uninsured drivers, crime suspects, and more. The LPR system plays a significant role in saving time and labor for institutions such as the police force. In the past, LPR relied heavily on Optical Character Recognition (OCR), which has been widely explored to recognize characters in images for numerous use cases. Usually, collected plate images suffer from various limitations and distortions, including noise, blurring, weather conditions, and close characters, making the recognition problem complex. Existing LPR methods still require significant improvement, especially for distorted images. To fill this gap, we propose utilizing visual language models (VLMs) such as OpenAI GPT4o (Generative Pre-trained Transformer 4 Omni), Google Gemini 1.5, Google PaliGemma (Pathways Language and Image model + Gemma model), Meta Llama (Large Language Model Meta AI) 3.2, Anthropic Claude 3.5 Sonnet, LLaVA (Large Language and Vision Assistant), NVIDIA VILA (Visual Language), and moondream2 to recognize such unclear plates with close characters. This paper evaluates the VLM's capability to address the aforementioned problems. Additionally, we introduce ``VehiclePaliGemma'', a fine-tuned Open-sourced PaliGemma VLM designed to recognize plates under challenging conditions. We compared our proposed VehiclePaliGemma with state-of-the-art methods and other VLMs using a dataset of Malaysian license plates collected under complex conditions. The results indicate that VehiclePaliGemma achieved superior performance with an accuracy of 87.6\%. Moreover, it is able to predict the car's plate at a speed of 7 frames per second using A100-80GB GPU. Finally, we explored the multitasking capability of VehiclePaliGemma model to accurately identify plates containing multiple cars of various models and colors, with plates positioned and oriented in different directions.
\end{abstract}

% \begin{figure}[t]
%     \centering
%     \includegraphics[width=0.9\linewidth]{images/blockDiagram.pdf}
%      % \includegraphics[width=0.8\linewidth]{images/vehicleGPT.pdf}
% \end{figure}

% \begin{highlights}
% \item We unveiled the potential of utilizing visual language models (VLMs), large and small, such as OpenAI GPT4o, Google Gemini 1.5,  Google PaliGemma, Meta Llama 3.2, Anthropic Claude 3.5 Sonnet,  LLaVA-NeXT, NVIDIA VILA, and moondream2, for unclear and close characters license plate recognition applications. 
% \item We evaluated the proposed solution and compared it with state-of-the-art deep OCRs using a dataset that contains Malaysian license plates collected in real life under complex conditions. The results show that VLMs are superior to traditional techniques, achieving an accuracy of 87.66\%.
% \item We proposed VehicleGPT (a multitasking GPT4) and VehiclePaliGemma (a multitasking PaliGemma) for localizing and recognizing plates' characters in images of multiple cars using a prompt engineered for a car with a specific color and model. 
% \end{highlights}

%% Keywords

\end{frontmatter}



\section{Introduction}
License plate recognition (LPR) systems, also known as automatic number plate recognition (ANPR), utilize optical character recognition on images to read vehicle registration plates. This widely recognized technique is instrumental in traffic management systems and has heaped significant focus on itself due to its real-time applications~\cite{anagnostopoulos2008license}. An advanced LPR system not only effectively recognizes car plates but also contributes significantly to improving traffic efficiency by distinguishing different classes of vehicles~\cite{lubna2021automatic}. The adoption of LPR systems in various areas has been growing over the years due to their wide-ranging benefits~\cite{wijersimplementing}. In law enforcement, for instance, LPR systems are employed to monitor traffic compliance, find stolen vehicles, and manage access control~\cite{du2012automatic}. In the area of toll systems, car plate recognition enables automatic toll collection, reducing congestion at toll booths. In parking management, ANPR reduces the need for manual ticketing and enables the efficient tracking of vehicles~\cite{kamaruzaman2021parkey}.

Despite the importance of this LPR system, there are a few limitations that still pose challenges. The advanced LPR system should be able to handle real-world conditions such as low illumination and weather changes (e.g., rain and snow). Additionally, the recognition system should be able to adapt to various other real-life limitations, such as the usage of low-quality cameras, unclear car plates, and complex backgrounds~\cite{idrose2022evaluation}.

The historical evolution of car plate recognition systems showcases a fascinating trajectory of technological advancements aimed at enhancing accuracy, speed, and adaptability. The inception of these systems can be traced back to the use of optical character recognition (OCR)-based approaches, which marked the early efforts to automate the extraction of textual information from vehicle registration plates~\cite{sugiyono2023extracting}. These early methods relied heavily on image processing techniques to detect, segment, and recognize characters on the plates, offering a foundational step towards automation. As technology progressed, the field witnessed significant enhancements with the integration of traditional machine-learning techniques~\cite{du2012automatic}. These algorithms, including support vector machines (SVMs) and neural networks, offered more robust feature extraction and classification methods, considerably improving the recognition rates under varied and challenging conditions. This era of car plate recognition was characterized by the deliberate shift from rule-based processing to data-driven approaches, enabling systems to learn from examples rather than follow explicitly programmed instructions~\cite{bishop2006pattern}.

Language models are fundamental elements of natural language processing (NLP). They predict the likelihood of a sentence by computing the probability distribution of the next word in the sentence given the words already seen~\cite{bengio2000neural}. With developments in deep learning, language models have begun to handle complex tasks in various sectors. In healthcare, for instance, language models help to improve healthcare delivery by analyzing electronic health records~\cite{yang2022large}. Similarly, in the education sector, language models are used to develop intelligent tutoring systems~\cite{modran2024llm}.

Parallel to the advancements of car plate recognition systems, the domain of NLP saw the introduction of large language models (LLMs)~\cite{vaswani2017attention,devlin2018bert}. These models, powered by deep learning architectures, have revolutionized the way machines understand human language. LLMs, such as the generative pre-trained transformer (GPT) by OpenAI~\cite{Hello_GPT-4o} and bidirectional encoder representations from transformers (BERT) by Google~\cite{devlin2018bert}, exhibit an unprecedented capacity to generate coherent text, comprehend context, and perform language understanding tasks with remarkable accuracy. The general capabilities of LLMs extend beyond text generation to include language translation, question answering, and text summarization, showcasing their versatility across various fields.

Pushing the boundaries of AI capabilities, visual language models (VLMs) are built upon the foundational work done in LLMs. VLMs are designed to process and understand both visual and textual data simultaneously. For instance, VLMs can generate descriptive texts from images, which could then be parsed for relevant information, including car plate data, effectively bridging the gap between visual data and language~\cite{radford2021learning}.

Exploring the potential of VLMs in car plate recognition systems presents an innovative research direction. The integration of VLMs could address some of the limitations of traditional methods, such as the handling of obscured or distorted plates and the adaptation to new plate formats without extensive retraining. The rationale behind leveraging VLMs lies in their ability to understand and interpret context, which could be beneficial in deciphering partially visible or damaged plates. Furthermore, their adaptability and generative capabilities suggest potential benefits in terms of accuracy and robustness, making them a promising tool in the continual evolution of car plate recognition technologies.

In this study, our proposed license plate recognition system utilizes state-of-the-art visual language models such as GPT4o~\cite{Hello_GPT-4o,GPT-4o}, Google's Gemini 1.5~\cite{Gemini_15_technical_report,Introducing_Gemini_15}, Google PaliGemma~\cite{beyer2024paligemma} , Meta Llama 3.2~\cite{Llama_vision}, Anthropic Claude 3.5 Sonnet~\cite{Claude_3.5_Sonnet}, LLaVA-NeXT~\cite{LLaVA,LLaVA-NeXT,Visual_and_Language}, VILA~\cite{{lin2024vila}}, and moondream2~\cite{moondream_1,moondream_2} to recognize plate's characters that are too close to each other and were captured under various challenging conditions. Our contributions can be summarized as follows:

\begin{enumerate}
    \item We explored the OCR capability of visual language models and employed them in the task of license plate recognition.
    \item We evaluated state-of-the-art visual language models such as GPT4o, Google Gemini 1.5, Google PaliGemma, Meta Llama 3.2, Anthropic Claude 3.5 Sonnet, LLaVA-NeXT, VILA, and moondream2 in terms of plate-level recognition accuracy and character-level accuracy.
    \item We utilized an image dataset of plates that were collected in real-life under various challenging conditions, including low illumination, low-quality cameras, unclear car plates, and close characters.
    \item We proposed two  multitasking VLMs, namely ``VehicleGPT'' and ``VehiclePaliGemma'' for localizing and recognizing plates' characters from images of multiple cars using a prompt engineered for a car with a specific color and modal. 
\end{enumerate}

The rest of the paper is organized as follows: In Section~\ref{relatedwork}, we review previous works on OCR and LPR. Section~\ref{researchMotivation} presents our research motivation. In Section~\ref{methods}, we describe the plate images collected to run the experiments and the methodology used by our LPR system. Section~\ref{results} discusses the experimental results and compares the proposed solution with other baseline methods. Finally, conclusions and future works are discussed in Section~\ref{conclusion}.

\section{Related Work}
\label{relatedwork}

\subsection {Traditional Methods of Car Plate Recognition}
Before the widespread application of deep learning techniques, car plate recognition systems largely hinged on optical character recognition (OCR) and traditional machine learning methods such as SVMs and k-nearest neighbor (KNN) models~\cite{anagnostopoulos2008license,gunawan2019automatic}. These technologies are aimed at identifying and classifying the characters of the license plates from the images. OCR methods were pivotal in converting different styles of vehicle number plate fonts into machine-encoded text. Machine learning methods like SVMs excelled at classifying segmented characters into recognizable letters and digits based on feature extraction from the input images~\cite{du2012automatic}.

Edge detection methods, such as the Canny edge detector~\cite{mousa2012canny}, have been widely used for identifying car parts in images by highlighting significant transitions in intensity. Similarly, color analysis techniques, such as histogram-based methods, are employed to distinguish cars from the background based on their color distribution~\cite{aruna2024detection}.

Template matching, which is another traditional method, involves comparing portions of the image with pre-defined templates of car shapes. Although this is useful in specific scenarios, template matching is computationally intensive and less adaptable to diverse real-world conditions~\cite{du2012automatic}.

Despite their successes, traditional methods faced notable limitations. The accuracy of these systems significantly declined in suboptimal conditions such as poor lighting, varied angles, motion blur, and diverse plate formats. These methods also struggled with the generalization needed to cope with the worldwide variety of license plate designs, requiring considerable manual tuning to adapt to each new format~\cite{wijersimplementing}.

\subsection {Deep Learning Approaches}
The advent of deep learning has significantly transformed car plate recognition systems, offering enhanced accuracy and robustness. The emergence of Convolutional
Neural Networks (CNNs) has substantially advanced the field of image recognition ~\cite{krizhevsky2012imagenet}. CNNs have been instrumental due to their hierarchical feature extraction capabilities, which accurately identified salient features in images without the need for manual feature design~\cite{lecun2015deep}. In the realm of car plate recognition, CNNs have demonstrated superior performance in detecting and recognizing number plates under various challenging conditions, outperforming traditional machine learning methods~\cite{montazzolli2017real}.

Several notable studies have emphasized the efficacy of CNNs in this domain. For instance, researchers developed a system employing CNNs that achieved remarkable accuracy in recognizing Brazilian car plates using two (You Only Look Once) YOLO-CNNs~\cite{montazzolli2017real}. This success underscores the CNNs potential to drastically mitigate the previous limitations through their adeptness at learning complex, variable patterns in data.

AlexNet~\cite{krizhevsky2012imagenet}, a pioneering CNN architecture, demonstrated the potential of deep learning in large-scale image classification tasks, setting the stage for its application in car plate recognition~\cite{he2022license}. Subsequent architectures like VGGNet~\cite{simonyan2014very} and ResNet~\cite{he2016deep} further improved the recognition performance by introducing deeper and more complex network structures~\cite{he2022license}.

Region-based CNNs (R-CNNs)~\cite{girshick2014rich} and their variants, such as Fast R-CNN~\cite{girshick2015fast} and Faster R-CNN~\cite{ren2015faster}, have been specifically tailored for object detection tasks, making them highly effective in identifying and localizing cars in images~\cite{saidani2021vehicle}. These models use region proposal networks to suggest potential bounding boxes, which are then refined by the CNN.

The YOLO family of models~\cite{redmon2016you,redmon2018yolov3}, known for their real-time detection capabilities, have also been applied to car plate recognition with impressive results~\cite{hendryli2020automatic}. YOLO's unified architecture, which performs detection and classification in a single forward pass, offers a balance between speed and accuracy.

More recently, transformers, originally designed for natural language processing, have been adapted for image recognition tasks. The Vision Transformer (ViT)~\cite{dosovitskiy2020image} leverages self-attention mechanisms to capture the global context in images, showing promise in car plate recognition applications~\cite{zhang2023automatic}.

\subsection {Emerging Use of LLMs in Image Processing}
The application of Large Language Models (LLMs) like GPT~\cite{{Hello_GPT-4o}} and BERT~\cite{devlin2018bert} transcends the barriers of text processing, venturing into non-text-based tasks including image recognition and processing. This expansion has been facilitated by the models' ability to understand and generate human-like text, providing a novel approach to interpreting and analyzing images~\cite{radford2021learning}.

Recent interdisciplinary studies have begun to explore the feasibility of LLMs for image-related tasks. For example, researchers have demonstrated the capabilities of GPT in generating textual descriptions from images, opening new pathways for image understanding and processing through natural language descriptions~\cite{radford2021learning}. 

Large language models (LLMs), like GPT and its successors, have primarily been recognized for their prowess in natural language understanding and generation. However, recent research has begun exploring their potential in image recognition tasks, often through multimodal learning approaches~\cite{abdelhamed2024you}. The integration of LLMs with car plate recognition systems is a nascent area of exploration that holds the potential to redefine the efficiencies of these systems.

Multimodal models, such as CLIP (Contrastive Language-Image Pretraining)~\cite{radford2021learning}, combine the strengths of LLMs and CNNs by training on pairs of images and their textual descriptions. CLIP has demonstrated state-of-the-art performance on a variety of image recognition benchmarks, including car plate recognition~\cite{radford2021learning}. By leveraging large-scale datasets of images and text, CLIP learns a joint representation space, enabling robust recognition even in zero-shot scenarios.

DALL-E~\cite{pmlr-v139-ramesh21a}, another multimodal model, generates images from textual descriptions, showcasing the potential of LLMs in understanding and creating visual content~\cite{pmlr-v139-ramesh21a}. While primarily a generative model, the principles underlying DALL-E's training could inform the development of more sophisticated car plate recognition systems.

The integration of LLMs with traditional vision models has also been explored through techniques like visual question answering (VQA)~\cite{antol2015vqa}, where models are trained to answer questions about images. These systems require a deep understanding of visual and textual information, highlighting the synergy between LLMs and image recognition~\cite{antol2015vqa}.

Recent work utilized three pre-trained OCR models, namely Tesseract~\cite{Tesseract_documentation}, EasyOCR~\cite{EasyOCR}, and KerasOCR~\cite{keras-ocr} and evaluated their performance in recognizing characters in complex car plates~\cite{idrose2022evaluation}. These models failed to recognize the characters in plate images under challenging conditions and produced low recognition accuracy~\cite{idrose2022evaluation}.

Our solution of utilizing VLMs for car plate recognition is proposed to address recognition problems under challenging conditions such as close characters and unclear plates and to improve the recognition accuracy largely using textual and visual understanding, as well as the OCR capability of VLMs for this purpose.

\section {Research Motivation}
\label{researchMotivation}

Although direct applications of VLMs in car plate recognition have yet to be extensively documented, the principles of the case studies---mentioned earlier in the related work section---offer intriguing prospects. The adaptability and contextual understanding of VLMs could potentially address complex challenges in car plate recognition, such as deciphering obscured or damaged plates and recognizing plates from diverse global formats without extensive reprogramming for each new case.

The insights from these studies suggest that VLMs, with their deep understanding and generation capabilities, could offer complementary, if not substitutive, solutions to traditional and CNN-based approaches in car plate recognition systems. By leveraging the advanced language comprehension and contextual analytics of VLMs, researchers could pave the way for breakthroughs in accuracy, efficiency, and adaptability in car plate recognition technologies.

\section{Materials and Methods}
\label{methods}

\subsection{Dataset Overview}
\subsubsection{Complex Plate Dateset}
\begin{figure}[hbt]
    \centering
    \includegraphics[width=1\linewidth]{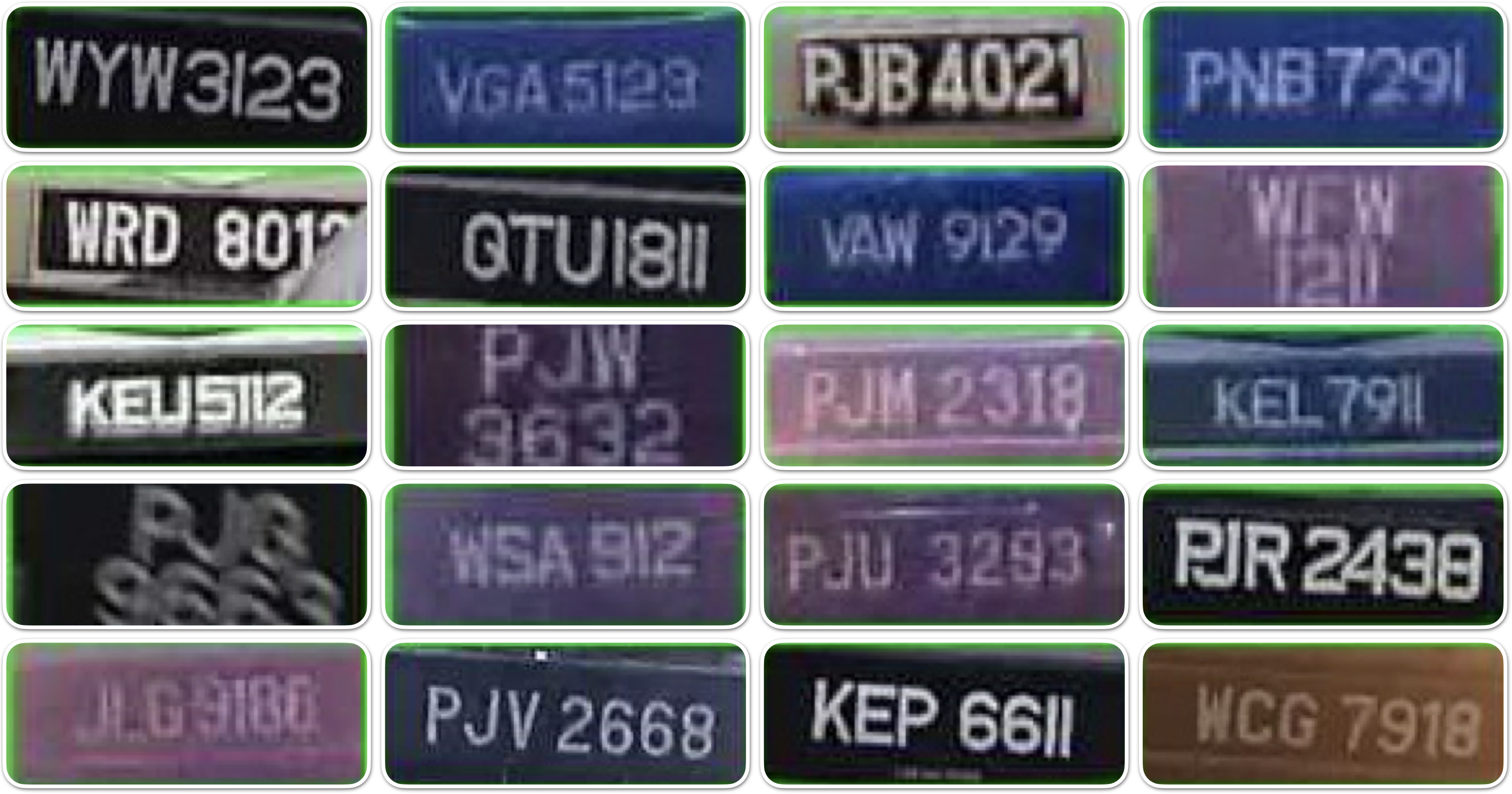}
    \caption{Sample complex license plates from the used dataset.}
    \label{fig:dataset}
\end{figure}

The license plate dataset used in this work consists of 258 labeled images of Malaysian license plates that are blurry, not clear, and have close characters. The dataset was collected by a Malaysian company called Tapway Sdn Bhd~\cite{VehicleTrack}. These images were considered complex and difficult to recognize by state-of-the-art OCR methods. Figure~\ref{fig:dataset} shows examples of these plates. 

This set of 258 images was collected for evaluation purposes only to test if the proposed solution is able to address the previous limitations and recognize the plates correctly (i.e., the gold set). Our researchers manually labeled the images to identify the characters in each one. This process was repeated three times, involving three different individuals, to ensure data consistency and accuracy. The final labels were determined using a voting technique to confirm the correct characters. The plate images have a width range of 64 to 181 pixels and a height range of 24 to 72 pixels. 

\subsubsection{Training Plate Dateset}
We developed a synthetic image dataset to fine-tune PaliGemma. This dataset comprises 600 images of Malaysian license plates, created with a black background and white alphanumeric characters (letters and numbers). Each image has a resolution of 50x120 pixels. Two plate formats were generated: a single line containing three letters followed by four numbers, and a two-line format where the first line includes three letters, and the second line contains four numbers. The letters and numbers were selected randomly. The images were rotated by 5 degrees in both directions, blurred, and subjected to Gaussian and salt-and-pepper noise.

\subsubsection{Diverse Car Dateset}
We scraped a dataset consisting of 140 images of single or multiple cars from the web with the key word ``Malaysian car plates''. We labeled these images by three evaluators with a majority voting technique as follows: if at least two evaluators, out of the three, gave the same label to the character, then this label is deemed to be correct. Otherwise, the character is checked again to have an agreement from at least two evaluators. This dataset was utilized to evaluate the multitasking capability of VehicleGPT and VehiclePaliGemma.

\begin{figure}[htb]
    \centering
    \includegraphics[width=0.8\linewidth]{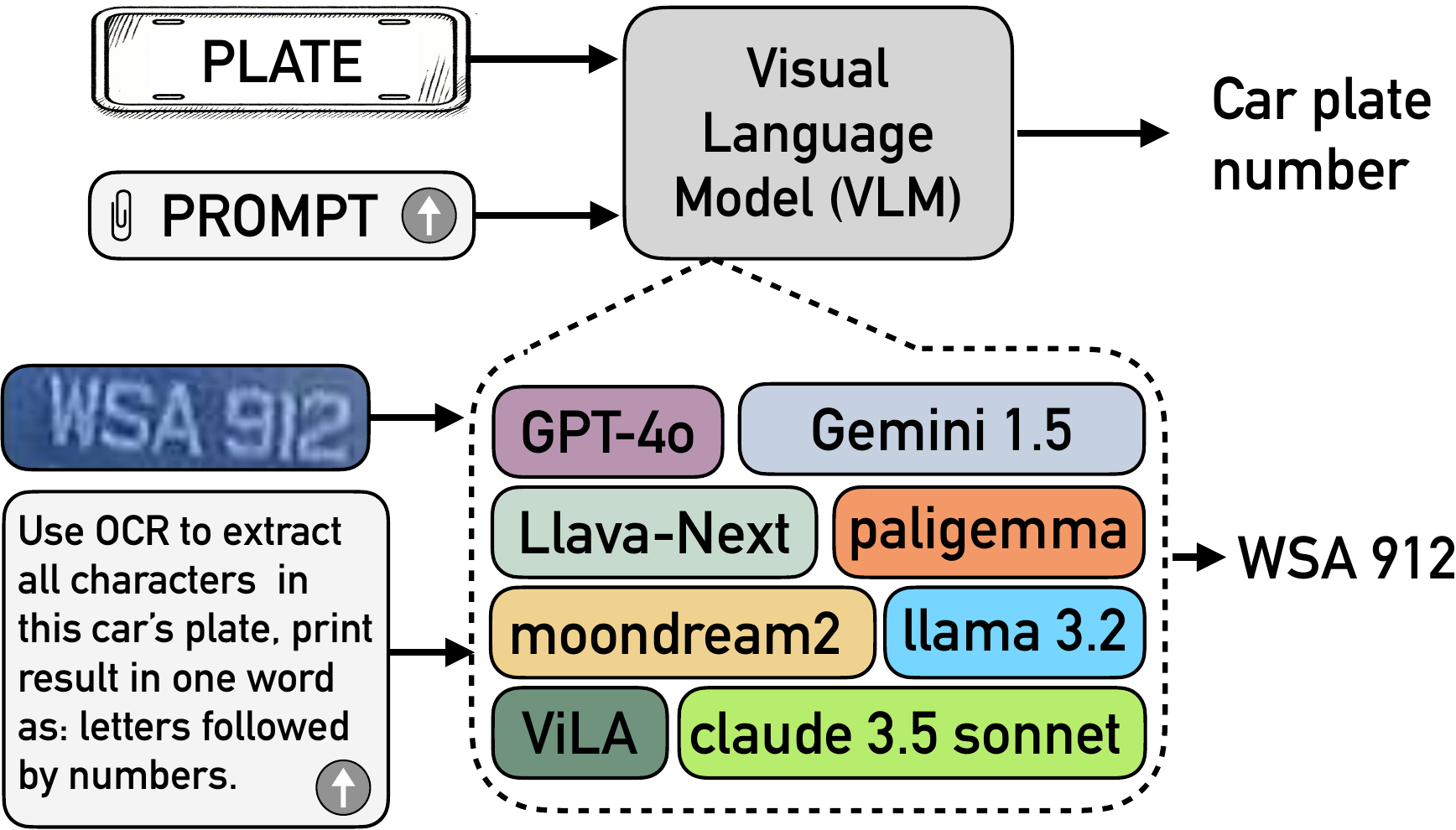}
    \caption{The proposed solution block diagram.}
    \label{fig:block_diagram}
\end{figure}

\subsection{Methods}
The proposed solution for license plate recognition is an artificial intelligence system that combines both language and visual processing to provide an enhanced understanding and generation capabilities and to extract characters from  car plate images given a proper prompt. We employed VLMs to utilize their natural language processing capabilities to interpret and analyze the context within the images. The solution utilizes the OCR capability of VLMs to understand the text, including the characters in the license plate, directly from the plate images without any preprocessing. Figure~\ref{fig:block_diagram} shows the block diagram of the proposed solution. 

As shown in Figure~\ref{fig:block_diagram}, the license plate image and text (i.e., the prompt) are applied to the inputs of each VLM, namely GPT4o~\cite{Hello_GPT-4o,GPT-4o}, Google's Gemini 1.5~\cite{Gemini_15_technical_report,Introducing_Gemini_15}, Google PaliGemma~\cite{beyer2024paligemma} , Meta Llama 3.2~\cite{Llama_vision}, Anthropic Claude 3.5 Sonnet~\cite{Claude_3.5_Sonnet}, LLaVA-NeXT~\cite{LLaVA,LLaVA-NeXT,Visual_and_Language}, VILA~\cite{{lin2024vila}}, and moondream2~\cite{moondream_1,moondream_2}.

We evaluated each of these VLMs separately and compared their outcomes against the ground truth. These VLMs represent the well-known VLMs available in the literature in both small- and large-size models. 

Each of these VLMs has the OCR capability to understand the contents of the image, such as their characters, and the language processing capability to understand the prompt given to the VLM asking them to perform a specific action on the given image. The VLM processes the plate image to recognize its characters and also uses its contextual understanding to ensure that the extracted text makes sense and aligns with the prompt's requirements. In this work, various VLMs such as GPT4o, Google's Gemini 1.5 Pro,  Meta Llama 3.2 11b, Anthropic Claude 3.5 Sonnet, LLaVA-NeXT-34b have been evaluated and compared to find the best model that can produce the highest recognition accuracy. Additionally, we also evaluated the performance of small vision language models (such as GPT-4o-mini, Gemini 1.5 Flash, Google PaliGemma 3b, LLaVA-NeXT-7b, VILA, and moondream2), which are designed to run efficiently on laptops or edge devices.
In this section, a summary of each used VLM is presented. The prompt that was used for the comparison is ``\texttt{Extract three letters and four numbers from this car's plate; print the result in one word as: letters followed by\\numbers}''.

\subsubsection{OpenAI Generative Pre-trained Transformer 4 Omni}

The Generative Pre-trained Transformer 4 Omni (GPT-4o)~\cite{GPT-4o,Hello_GPT-4o} is the first VLM used in this study. It has vision capabilities and is a big step forward in AI because it combines powerful language processing with complex image analysis. This multimodal model integrates visual understanding with textual analysis, expanding the functionality of AI applications. GPT-4o excels in visual question answering (VQA), allowing users to input images alongside questions to receive contextually relevant answers. Additionally, GPT-4o demonstrates strong optical character recognition (OCR) capabilities, effectively extracting and interpreting text from images, which benefits document digitization and reading signs in images~\cite{GPT-4o,Hello_GPT-4o}. The model's ability to combine image and text processing enables comprehensive and nuanced responses. For example, GPT-4o can describe image contents, generate captions, or analyze charts and graphs for insights. Its improved contextual understanding enhances its utility in continuous engagement applications~\cite{GPT-4o,Hello_GPT-4o}. Additionally, we used GPT-4o mini, which is the most advanced model in the small models category~\cite{Hello_GPT-4o}. It is the cheapest, most affordable, and most intelligent small model for fast and lightweight multimodal tasks (accepting text or image inputs and outputting text).

\subsubsection{Google Gemini-1.5}
The second VLM utilized in this work is Google Gemini-1.5~\cite{Introducing_Gemini_15}. This paper explored two versions of Gemini-1.5: the large Gemini 1.5 Pro and the small Gemini 1.5 Flash.
The Gemini 1.5 Pro is a mid-size multimodal model optimized for a wide range of tasks~\cite{Introducing_Gemini_15}. It features a context window of up to one million tokens, enabling it to seamlessly analyze, classify, and summarize large amounts of content within a given prompt. When compared to the largest 1.0 Ultra model~\cite{Introducing_Gemini_15,Gemini_15_technical_report} on the same benchmarks, it performs at a broadly similar level. Additionally, Gemini 1.5 Pro demonstrates impressive in-context learning abilities, allowing it to acquire new skills from information provided in a long prompt without requiring additional fine-tuning.

On the other hand, Gemini 1.5 Flash~\cite{Introducing_Gemini_15,Gemini_15_technical_report} represents a significant leap in AI technology by integrating multimodal capabilities with an emphasis on speed and efficiency. This model is designed to handle high-frequency tasks at scale, making it ideal for applications requiring rapid, real-time processing of both text and visual data. One of the standout features of Gemini-1.5 Flash is its long context window, which can process up to one million tokens~\cite{Introducing_Gemini_15,Gemini_15_technical_report}. In terms of strengths, Gemini-1.5 Flash excels in multimodal reasoning, effectively integrating text and visual information to deliver accurate and insightful outputs. Its efficiency is bolstered by a streamlined architecture using a ``distillation'' process, where essential knowledge from larger models is transferred to this smaller, more efficient model. This makes it highly cost-effective and accessible for a wide range of users, from developers to enterprise customers.

\subsubsection{Google PaliGemma}

Google's PaliGemma is an open vision-language model (VLM) that extends the PaLI series by integrating it with the Gemma family of language models. Built upon the SigLIP-So400m vision encoder and the Gemma2b language model, PaliGemma serves as a versatile and broadly applicable base model, excelling in transfer learning~\cite{beyer2024paligemma}. It showcases strong performance across diverse open-world tasks, leveraging multi-task learning through task prefixes. The prefix-LM approach, which uses task prefixes and supervises only suffix tokens, proves to be an effective pre-training objective for VLMs.

While fine-tuning is useful for solving specific tasks, a generalist model with a conversational interface is often preferred. Instruction tuning, achieved by fine-tuning on a diverse dataset, typically facilitates this versatility. PaliGemma has been shown to be well-suited for such transfer learning~\cite{beyer2024paligemma}.

In this work, we employed two versions of PaliGemma: the pre-trained PaliGemma and a fine-tuned version named VehiclePaliGemma, specifically optimized for the car's license plate recognition task. The VehiclePaliGemma was fine-tuned using training plate dataset (see the Dataset section above). The hyper-parameters utilized for fine-tuning are: learning rate with 0.00002, train batch size with 2,
gradient accumulation steps with 8, Adam optimizer, and five epochs. 
The outcome of the fine-tuning was fine-tuned PaliGemma, ``VehiclePaliGemma'', that we open-sourced on the Hugging Face platform \url{https://huggingface.co/NYUAD-ComNets/VehiclePaliGemma}

\subsubsection{Llama Instruct}

Llama 3.1, developed by Meta, is an auto-regressive language model built on an optimized transformer architecture~\cite{Llama}. It includes multilingual LLMs that offer both pre-trained and instruction-tuned generative models, designed to handle text inputs and outputs effectively. 

Llama 3.2 Instruct with vision capability ~\cite{Llama_vision} extends the Llama 3.1 text-only model into a multi-modal generative framework capable of processing both text and image inputs to generate text outputs. Optimized for tasks like visual recognition, image reasoning, captioning, and answering questions about images, Llama 3.2 Instruct employs instruction tuning. It integrates a separately trained vision adapter to handle image recognition, which works in conjunction with the pre-trained Llama 3.1 language model. In this study, we evaluated Llama 3.2 11b model to support our efforts in recognizing complex car's plate by combining object recognition in images with semantic analysis of text.

\subsubsection{Claude 3.5 Sonnet}

Claude 3.5 Sonnet establishes new industry standards~\cite{Claude_3.5_Sonnet}. It demonstrates significant advancements in understanding nuance, humor, and intricate instructions, excelling at producing high-quality content with a natural and relatable tone. Operating at twice the speed of Claude 3 Opus, Claude 3.5 Sonnet delivers a substantial performance boost. Its enhanced efficiency, paired with cost-effective pricing, makes it an excellent choice for complex tasks.

Claude 3.5 Sonnet is the most advanced Anthropic vision model to date, outperforming Claude 3 Opus on standard vision benchmarks. Its significant enhancements are particularly evident in tasks requiring visual reasoning, such as analyzing charts and graphs. Additionally, Claude 3.5 Sonnet excels at accurately transcribing text from imperfect images—a critical capability for industries like retail, logistics, and financial services. In this work, we explored and evaluated the capability of Claude 3.5 Sonnet model to recognize complex car's plates.

\subsubsection{LLaVA-NeXT}
The third VLM demonstrated in this work is Large Language and Vision Assistant (LLaVA)~\cite{LLaVA}. LLaVA-NeXT~\cite{LLaVA-NeXT} represents a significant advancement in multimodal AI models, designed to integrate and enhance both language and vision capabilities. This model is built upon the success of its predecessor, LLaVA, incorporating improvements in reasoning, optical character recognition (OCR), and overall world knowledge. LLaVA-NeXT excels in visual question answering (VQA) and image captioning, leveraging a combination of a pre-trained large language model (LLM) and a vision encoder. The model's architecture enables it to handle high-resolution images dynamically, preserving intricate details that improve visual understanding~\cite{LLaVA,LLaVA-NeXT,Visual_and_Language}. The model's efficiency is another key strength. LLaVA-NeXT achieves state-of-the-art performance with relatively low training costs, utilizing a cost-effective training method that leverages open resources~\cite{LLaVA-NeXT}. Despite its strengths, LLaVA-NeXT faces challenges in handling extremely complex visual tasks that may require specialized models for optimal performance. Additionally, while it has shown strong results in zero-shot scenarios, further refinement is needed to consistently match or exceed the performance of commercial models in all contexts~\cite{LLaVA,LLaVA-NeXT,Visual_and_Language}. Several versions of LLaVA are available based on the number of parameters (i.e., the model's size). We utilized two versions in our experiments: large 34 billion LLaVA and small 7 billion LLaVA.

\subsubsection{Visual Language Model (VILA)}
It is notably worth considering the computational requirements of VLMs, which are usually important for the practical implementation of such systems in real-world scenarios~\cite{brown2020language}. Therefore, in this work, small versions of VLMs such as VILA~\cite{lin2024vila} have also been explored for plate recognition. VILA is a very recent VLM pre-trained with interleaved image-text data at scale, enabling multi-image VLM~\cite{lin2024vila}.	It unveils appealing capabilities, including multi-image reasoning, visual chain-of-thought, and video understanding. VILA was found to outperform state-of-the-art models like LLaVA-1.5 across various benchmarks. Furthermore, VILA is deployable on the edge via AWQ 4bit quantization. In this work, we utilized the Llama-3-VILA1.5-8B~\cite{lin2024vila} version to recognize characters in plate images.  

\subsubsection{moondream2}
Another VLM that is used in this work is moondream2~\cite{moondream_1,moondream_2}. It is an open-source tiny and compact visual language model incorporating weights from the Sigmoid Loss for Language Image Pre-Training (SigLIP) and Phi-1.5 small language models. moondream2 is specifically engineered for efficient operation on devices with limited computational capabilities, such as edge devices with very little memory~\cite{moondream_1,moondream_2}. 

\section{Results and Discussion}
\label{results}
This section presents the results of evaluating and comparing our proposed solution, which leverages the OCR capabilities of VLMs to address the challenging problem of car plate recognition. Several VLMs were evaluated and compared in terms of plate-level accuracy and character-level accuracy. Additionally, we compared the proposed solution with three pre-trained deep learning OCR models, namely Tesseract~\cite{Tesseract_documentation}, EasyOCR~\cite{EasyOCR}, and KerasOCR~\cite{keras-ocr}. The comparison was done using a complex plate dataset that contains complex Malaysian license plates (see the Dataset section above). 

We conducted several experiments to evaluate the vision capabilities of the VLMs for: 1) the OCR task in general, and 2) license plate recognition in particular. In the first experiment, we examined GPT-4's vision capabilities and employed OCR to extract characters from the plate images. Integrating OCR with GPT-4 allows the extracted text to be combined with the language model, enhancing the model's understanding and processing of both the image and any associated text. Table~\ref{tab:character_level} shows the character-level accuracy of GPT-4o (97.1\%) by recognizing 1700 correct characters out of 1751 characters. Similarly, the GPT-4o mini version gave a close accuracy of 96.7\%. Additionally, we investigated the Google Gemini 1.5 Pro model to study the OCR capability of Gemini for our plate recognition task. The results indicate degradation in character-level accuracy in both Gemini 1.5 Pro (93.8\%) and Gemini 1.5 Flash (93.8\%). Similarly, Llama 3.2 Instruct and Claude 3.5 Sonnet produced less recognition accuracy (93.38\% and 92.8\%, respectively) compared to Gemini 1.5. Likewise, LLaVA-NeXT has less recognition accuracy compared to the previously mentioned VLMs, producing a character-level accuracy of 85.9\% in the 34b version and 80.94\% in the 7b version. In contrast, small VLM versions such as VILA show better recognition performance than the LLaVA-NeXT 7b with accuracy of 83.21\%. Furthermore, the tiny moondream2 has less recognition capability than VILA with a character-level accuracy of 76.58\%. The results indicate that the two small versions of VLMs, namely GPT-4o mini and Gemini 1.5 Flash, outperformed other small VLMs such as VILA and moondream2 in our plate recognition task. The number of correctly predicted characters for each VLM is shown in Table~\ref{tab:character_level}.

Using the pre-trained PaliGemma model, a character-level accuracy of 90.92\% was achieved, correctly recognizing 1,592 characters out of 1,751. In contrast, the fine-tuned version, VehiclePaliGemma, demonstrated a significant improvement, increasing character-level accuracy by 7\% to reach 97.66\%, with 1,710 characters correctly identified. This performance surpasses other VLMs in general, including GPT-4o, as detailed in Table~\ref{tab:character_level}. 

\begin{table}[]
\centering
\begin{tabular}{|l|c|c|}
\hline
\textbf{Method}               & \textbf{Number of correctly} & \textbf{character-level} \\     
     & \textbf{predicted characters} & \textbf{Accuracy \%} \\ \hline

\textbf{moondream2}           & 1341                                              & 76.58 \%                        \\ \hline
\textbf{VILA}                 & 1457                                              & 83.21 \%                        \\ \hline
\textbf{LLaVA-NeXT-7b}        & 1417                                              & 80.93 \%                        \\ \hline
\textbf{Gemini 1.5   flash}   & 1643                                              & 93.8 \%                         \\ \hline
\textbf{GPT-4o-mini} & 1693                                     & 96.7 \%                \\ \hline
\textbf{LLaVA-NeXT-34b}       & 1504                                              & 85.9 \%                         \\ \hline
\textbf{Gemini 1.5 Pro}       & 1643                                              & 93.8 \%                         \\ \hline
\textbf{GPT-4o}        & 1700                                     & 97.1 \%                \\ \hline
\textbf{Llama 3.2 Instruct}        & 1635                                     & 93.38 \%                \\ \hline
\textbf{Claude 3.5 Sonnet}        & 1625                                     & 92.80 \%                \\ \hline

\textbf{Pre-trained PaliGemma}        & 1592                                     & 90.92 \%                \\ \hline

\textbf{VehiclePaliGemma}        & \textbf{1710}                                     & \textbf{97.66 \%}                \\ \hline

\end{tabular}
\caption{Character-level accuracy results of several VLMs.}
\label{tab:character_level}
\end{table}

In the second experiment, we compared our proposed solution of utilizing VLMs with state-of-the-art methods. Comparing traditional approaches with VLM-based methodologies reveals substantial differences in potential outcomes, as seen in Table~\ref{tab:plate_level}.
Three pre-trained deep learning models, namely KerasOCR, EasyOCR, and Tesseract, are considered baseline methods in this work and were used for comparison. These models that showed promising performance in various OCR tasks~\cite{tesseract_app,smelyakov2021effectiveness,vedhaviyassh2022comparative} failed to recognize the characters in plate images in our dataset~\cite{idrose2022evaluation}. Tesseract 4.0 is an OCR engine based on Long Term Short Memory (LSTM) neural networks~\cite{idrose2022evaluation}. EasyOCR detects Text using the Character-Region Awareness for Text detection (CRAFT) algorithm~\cite{idrose2022evaluation}. After that, EasyOCR utilizes Convolutional Recurrent Neural Network for recognition. Its recognition model contains several components: feature extraction (Resnet and VGG), sequence labelling (LSTM) and decoding (Connectionist Temporal Classification). KerasOCR utilizes CRAFT to detect text areas by analyzing each character region and the affinity between characters~\cite{idrose2022evaluation}. To locate text-bounding boxes, minimum-bounding rectangles are identified on the binary map after thresholding the scores of the character regions and their affinities. For text recognition, it employs either the original CRNN model or a spatial transformer network layer to rectify the text. 

\begin{table}[htb]
\centering
\begin{tabular}{|l|c|c|}
\hline
\textbf{Method}                 & \textbf{Number of correctly} & \textbf{Plate-level} \\
  & \textbf{predicted plates} & \textbf{Accuracy \%} \\ \hline
\textbf{EasyOCR (baseline)}     & 79                                            & 32.95\%                    \\ \hline
\textbf{Tesseract (baseline)} & 97                                            & 36.74\%                    \\ \hline
\textbf{KerasOCR (baseline)}    & 107                                           & 40.53\%                    \\ \hline
\textbf{Moondream2}             & 102                                           & 39.5 \%                    \\ \hline
\textbf{LLaVA-NeXT-7b}          & 144                                           & 55.8 \%                    \\ \hline
\textbf{VILA}                   & 147                                           & 57 \%                      \\ \hline
\textbf{Gemini 1.5   flash}     & 200                                           & 77.5 \%                    \\ \hline
\textbf{GPT-4o-mini}   & 220                                  & 85.7 \%           \\ \hline
\textbf{LLaVA-NeXT-34b}         & 152                                           & 58.9 \%                    \\ \hline
\textbf{Gemini 1.5 Pro}         & 185                                           & 71.7 \%                    \\ \hline
\textbf{GPT-4o}          & 222                                  & 86 \%             \\ \hline
\textbf{Llama 3.2}          & 175                                  & 67.83 \%             \\ \hline
\textbf{Claude 3.5 Sonnet}          & 186                                  & 72.1 \%             \\ \hline
\textbf{Pre-trained PaliGemma}          & 178                                 & 69 \%             \\ \hline
\textbf{VehiclePaliGemma}          & \textbf{226}                                  & \textbf{87.6 \%}             \\ \hline
\end{tabular}
\caption{Plate-level accuracy, comparing the performance of several VLMs against multiple baseline methods.}
\label{tab:plate_level}
\end{table}

The results in Table~\ref{tab:plate_level} show the plate-level accuracy and the number of correctly predicted plates. KerasOCR was able to recognize 107 images out of 258 images~\cite{idrose2022evaluation}, while EasyOCR and Tesseract predicted correctly 87 images and 97 images~\cite{idrose2022evaluation}, respectively. However, all of these three methods have low recognition accuracy and limitations that have been addressed in this work by leveraging the OCR capability of VLMs, as shown in Table~\ref{tab:plate_level}.

Among large pre-trained VLMs, GPT-4o achieved the highest plate accuracy at 86\%, correctly recognizing 222 out of 258 plates in the dataset. Claude 3.5 Sonnet ranked second with a plate accuracy of 72.1\%, followed by Gemini 1.5 Pro in third place at 71.7\%. VILA-NEXT 34b ranked last among them, achieving a plate accuracy of 58.9\%. On the other hand, among the small VLMs, GPT-4o mini achieved the highest plate accuracy at 85.7\%, followed by Gemini 1.5 flash with an accuracy of 77.5\%, outperforming its larger counterpart, Gemini 1.5 Pro. Pre-trained PaliGemma 3b secured third place with a plate accuracy of 69\%, while Llama 3.2 11b ranked fourth at 67.83\%.
Furthermore, other small VLMs such as VILA, LLaVA-NeXT, and moondream2 have accuracies of 57\%, 55.8\%, and 39.5\%, respectively. All small VLMs except moondream2 were able to outperform the three baseline methods.

The pre-trained PaliGemma model achieved a plate-level accuracy of 69\%, correctly recognizing 178 plates out of 258. In comparison, the fine-tuned version, VehiclePaliGemma, exhibited a substantial improvement, increasing plate-level accuracy by 18\% to 87.6\%, with 226 plates accurately identified. This performance notably exceeds that of other VLMs, including GPT-4o, as shown in Table~\ref{tab:plate_level}.

\begin{figure}[!htb]
    \centering
    \includegraphics[width=1\linewidth]{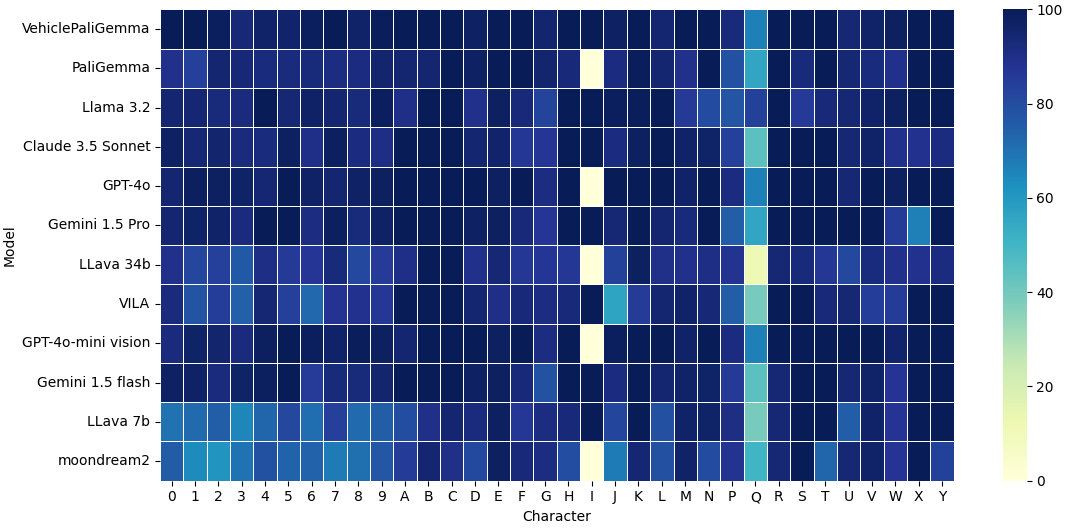}
    \caption{Character-level accuracy heatmaps for different vision models.}
    \label{fig:heatmap}
\end{figure}

The number of correctly predicted plates for each VLM utilized is shown in Table~\ref{tab:plate_level}. The heatmap of each character's accuracy for each VLM is shown in Figure~\ref{fig:heatmap}. The heatmap helps in quickly identifying which models perform consistently across all characters and which ones have variability in their recognition. The lighter colors indicate any particular characters where the models have struggled to identify them.

The results show that traditional systems relying on optical character recognition and machine learning face challenges in adaptability and require extensive manual tuning to maintain high accuracy under varied conditions. On the other hand, VLMs, with their sophisticated understanding of context and nuance, hypothetically promise greater adaptability and accuracy, especially in interpreting obscured or complex plate images. In the end, while VLMs offer a promising avenue for enhancing car plate recognition systems, their integration demands careful attention to computational feasibility and ethical standards. 

Integrating VLMs into such plate recognition systems requires careful consideration of ethical standards, as follows:
\begin{enumerate}
\item Ensuring that the deployment of these systems respects individuals' privacy, especially in public spaces where data might be collected without consent.
\item addressing any potential biases in the model that could lead to unfair treatment of certain groups, particularly in law enforcement contexts.
\item maintaining transparency in how these models make decisions and ensuring there is accountability for any errors or misuse.
\item safeguarding the data collected and used by these systems to prevent unauthorized access or misuse.
\item adhering to local and international laws regarding data collection, storage, and usage, particularly in relation to surveillance and data protection.
\end{enumerate}

\subsection{Prompt Sensitivity}
In this section, we studied the impact of prompts in VLMs on our plate recognition task. We chose three VLMs: VehiclePaliGemma, GPT-4o and Gemini 1.5 Pro due to their demonstrated superior performance in license plate recognition, as evidenced in prior results. We evaluated four prompts as follows:

\begin{itemize}
    \item Prompt1: ``\texttt{extract characters in this car's plate, print\\result in one word as: letters followed by numbers}''
    \item Prompt2: ``\texttt{extract three letters and four numbers from this\\car's plate; print the result in one word as: letters\\followed by numbers}''
    \item Prompt3: ``\texttt{use OCR to extract all characters in this car's\\plate, print result in one word as: letters followed by\\numbers}''
    \item Prompt4: ``\texttt{extract the text from the image}''
\end{itemize}

In the first prompt, we asked both GPT-4o and Gemini 1.5 Pro to extract characters in general without determining the number of letters and numbers in the license. On the other hand, prompt2 explicitly determined the exact number of letters and characters, i.e., four letters and three numbers, which can help in identifying all characters in the plates without missing any, thus increasing the number of correctly recognized plates as shown in Table~\ref{tab:prompt}. The previous advantages can be achieved only if all plates under evaluation have the same format (four letters followed by three numbers). Otherwise, the second prompt fails if we have plates with various formats. In the third prompt, we asked both GPT-4o and Gemini 1.5 Pro to use OCR to extract all characters, and the results in Table~\ref{tab:prompt} show the capability of GPT-4o to recognize 227 plates correctly out of 258 plates with an accuracy of 88\% using prompt3 which has more recognition capability when used in comparison to prompt2. In contrast, Gemini 1.5 Pro performed better with prompt2 compared to prompt3. Moreover, we evaluated VehiclePaliGemma with two prompts: prompt2 (that both GPT-4o and Gemini 1.6 Pro show good performance utilizing it) and prompt4. The plate accuracy with prompt4 was better than one with prompt2 by 23\%. The results show that VLMs are sensitive to prompts used to recognize characters in the plate images, and that careful attention should be given to the prompt to achieve the highest performance.  

To study the limitations of VLMs, we chose VehiclePaliGemma, which was the top recognition model in our experiments. First, we show the limitations using prompt4 as follows:

\begin{enumerate}
    \item Actual P is predicted as R, such as these pairs of examples (actual, predicted): (PJG90, RJG90), (PJW6633, RJW6633), (PJV8666, RJV8666), (PJC5688, RJC5688). It is clear in most cases that when J comes after P, the model predicts P as R, as shown in the following images.

    \begin{figure}[!htb]
        \centering
    \includegraphics[width=1\linewidth]{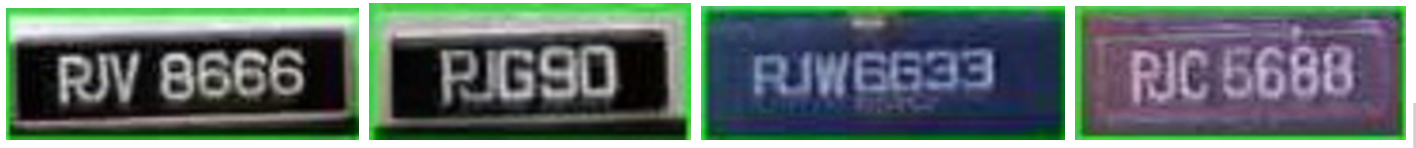}
    \end{figure}

    \item In few cases, When plates have only six characters (three letters and 3 numbers), VehiclePaliGemma added one letter, such as these pairs of examples (actual and predicted): (PJN214, PJN2114), (KCJ112,   KCJ1112), and (PLA113, PLA1113), as shown in the following images.

    \begin{figure}[!htb]
        \centering
    \includegraphics[width=1\linewidth]{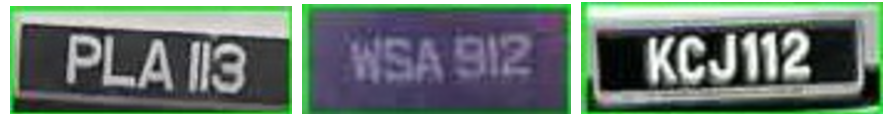}
    \end{figure}

    \item If a letter comes at the end, VehiclePaliGemma will reorder them according data fine-tuned on and put letters before numbers (actual: W1209G, predicted: WI2096).
\end{enumerate}

% \usepackage{multirow}
\begin{table}[!htb]
\centering
\begin{tabularx}{\columnwidth}{|X|c|c|c|}
\hline
\textbf{Method}                          &         & \textbf{Number of correctly} & \textbf{Plate Accuracy} \\ 
   &         & \textbf{predicted plates} & \textbf{(\%)} \\ \hline

{\textbf{GPT-4o}}  & Prompt1 & 216                                         & 83.7 \%                    \\ \cline{2-4} 
                                         & Prompt2 & 222                                         & 86 \%                      \\ \cline{2-4} 
                                         & Prompt3 & \textbf{227}                                & \textbf{88 \%}             \\ \hline
{\textbf{Gemini 1.5 Pro}} & Prompt1 & 177                                         & 68.6 \%                    \\ \cline{2-4} 
                                         & Prompt2 & \textbf{186}                                & \textbf{72.1 \%}           \\ \cline{2-4} 
                                         & Prompt3 & 176                                         & 68.2 \%                    \\ \hline
{\textbf{Pre-trained PaliGemma}} & Prompt2 & 119                                         & 46.12 \%                    \\ \cline{2-4} 
                                         & Prompt4 & \textbf{178}                   & \textbf{69} \%   
\\ \hline
\end{tabularx}
\caption{Prompt sensitivity in GPT-4o and Gemini 1.5 Pro.}
\label{tab:prompt}
\end{table}

The use of Visual Language Models (VLMs) for OCR in general, and specifically for license plate recognition, demonstrates significant potential for future applications that remain challenging for traditional machine learning models. Future advancements aimed at enhancing the visual analysis capabilities of VLMs could significantly increase their applicability for image analysis and understanding tasks, such as license plate recognition or any other complex use cases. However, to enhance their capabilities, more diverse and high-quality data are required to further improve the model's generalization capabilities.

\subsection{VehicleGPT and VehiclePaliGemma}
In this section, we propose ``VehicleGPT'' (a multitasking GPT-4o) and ``VehiclePaliGemma'' (a multitasking PaliGemma) with a car's plate recognition capability. It was able to detect (localize and recognize) cars' plates in images with single or multiple cars. We chose both LLMs due to theirs demonstrated superior performance in license plate recognition, as evidenced in prior results. 

\begin{figure}[!hbt]
    \centering
    \includegraphics[width=0.8\linewidth]{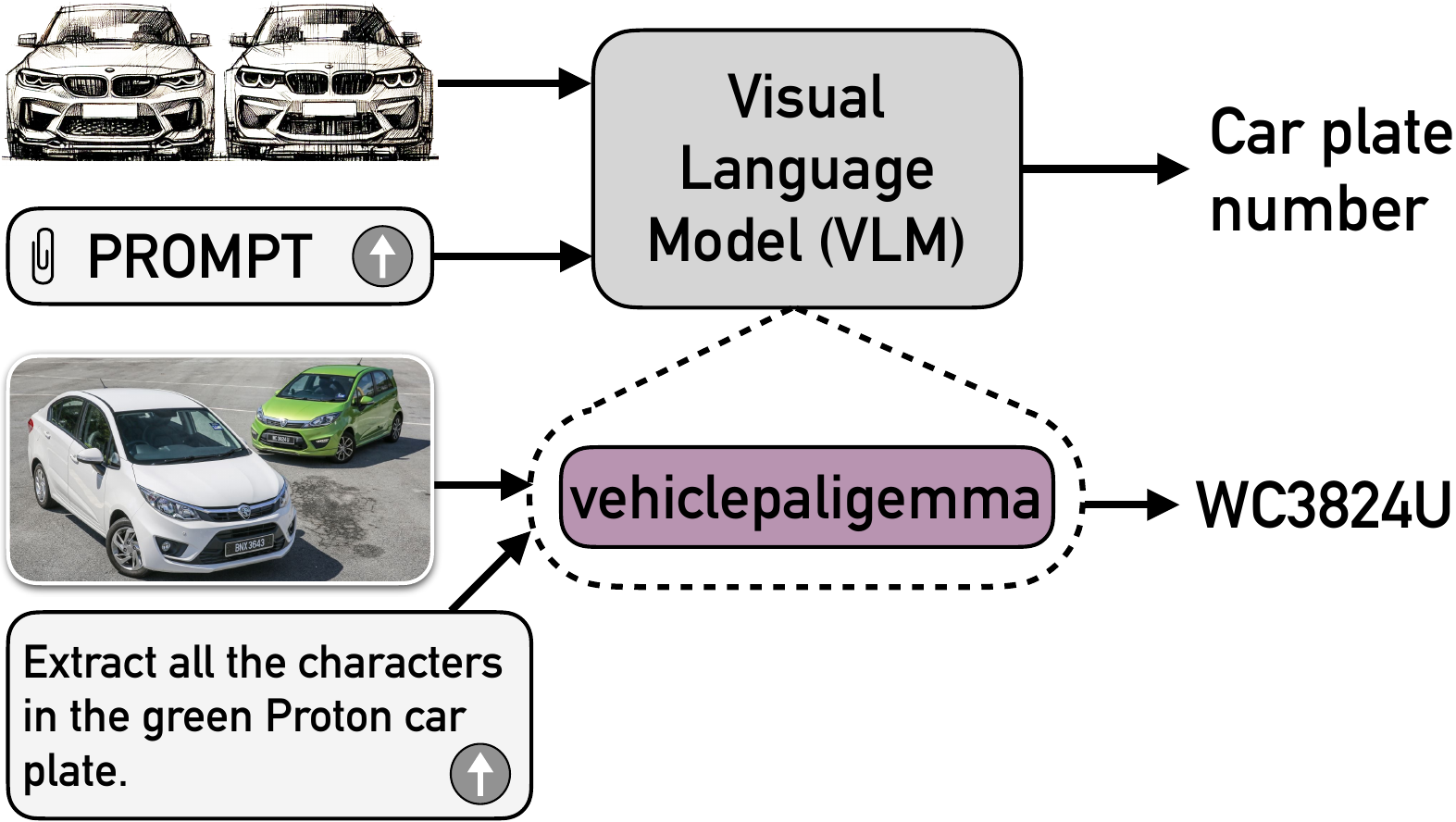}
    \caption{VehiclePaliGemma's block diagram.}
    \label{fig:VehicleGPT}
\end{figure}

\begin{figure}[!hbt]
    \centering
    \includegraphics[width=1\linewidth]{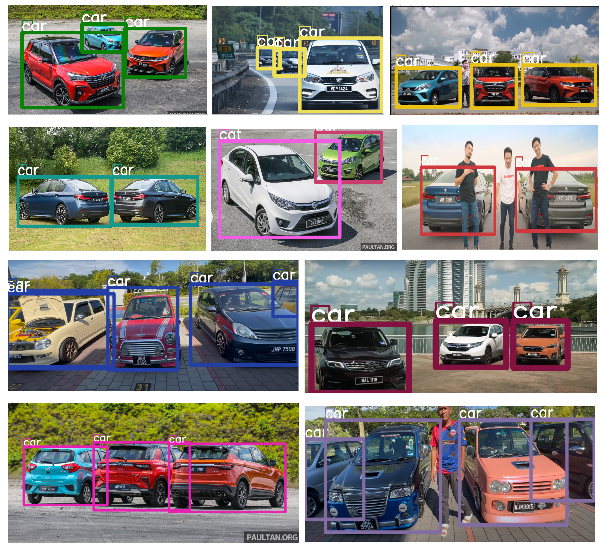}
    \caption{samples of cars detected by VehiclePaliGemma}
    \label{fig:paligemma_detection_car}
\end{figure}

\begin{figure}[!hbt]
    \centering
    \includegraphics[width=1\linewidth]{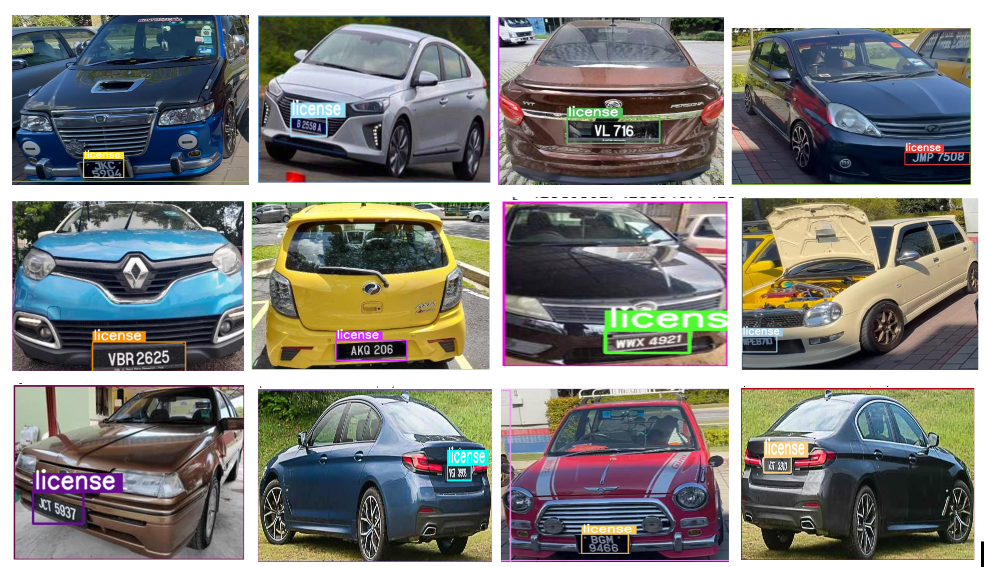}
    \caption{samples of license plates detected by VehiclePaliGemma}
    \label{fig:paligemma_plate_detection}
\end{figure}

Figure~\ref{fig:VehicleGPT} illustrates the block diagram of the proposed solution of VehiclePaliGemma. In this analysis, the input is an image of a single or multiple car(s), and the prompt used was ``\texttt{Extract all characters from the plate of the green Toyota car(s)}''. The output is the extracted characters from the specific car(s) referred to in the prompt. 
To detect cars and plates, and then recognize characters in the plates, our proposed solution VehiclePaliGemma followed several steps:

\begin{enumerate}
    \item using `\texttt{detect car}'' prompt to utilize the detection capability of pre-trained PaliGemma to localize all cars available in the images as shown in Figure~\ref{fig:paligemma_detection_car}.

    \item using `\texttt{detect license plate}'' prompt to leverage the pre-trained PaliGemma model's detection capabilities for localizing the plate of an already detected car as shown in Figure~\ref{fig:paligemma_plate_detection}.

    \item using `\texttt{extract the text from the image}'' prompt with VehiclePaliGemma to recognize characters in the detected plate.

     \item if the main prompt has a specific color or model of the car, pre-trained PaliGemma was asked to check the color and model before steps 2 and 3. For example, `\texttt{Is this car red/Toyota?}''.
    
\end{enumerate}

First, Both VehicleGPT and VehiclePaliGemma were evaluated with Diverse car dataset (see the Dataset section above) using``\texttt{Extract all characters from the car plates}'' prompt targeting all plates for all cars displayed in the image. The accuracy is calculated as follows: if the model recognizes all plates in the image correctly, the counter that counts the number of correctly recognized images is incremented by one. Otherwise, even if one plate in the image is not properly recognized, the counter is not incremented. The percentage of correctly identified images over the total number of images in the dataset determines the final accuracy. VehicleGPT identified successfully  171 plates among the 176 cars or plates present in 140 images, resulting in a plate-level accuracy of 97.16\%. Similarly, VehiclePaliGemma recognized correctly 166 plates, resulting in a plate-level accuracy of 94.32\%.

\begin{figure}[!hbt]
    \centering
    \includegraphics[width=1\linewidth]{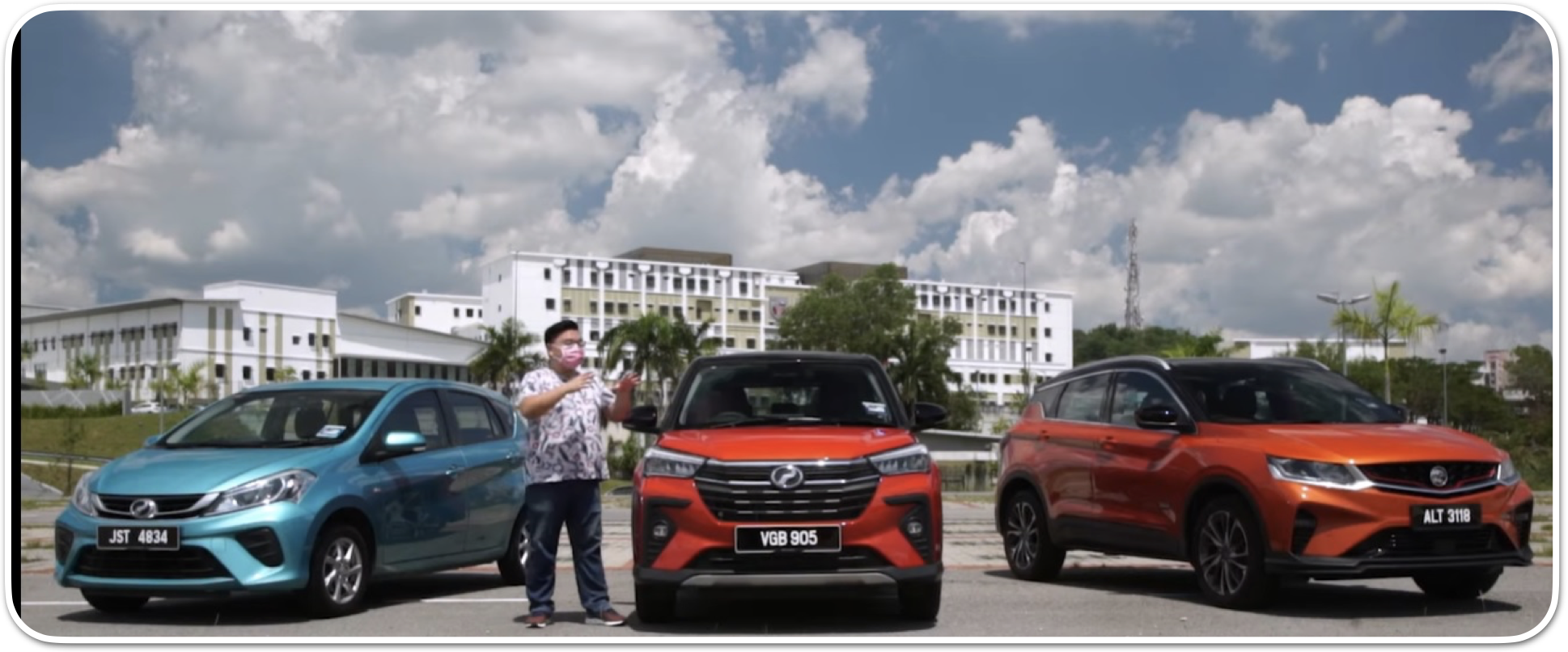}
    \caption{Example for prompt 1 featuring red cars.}
    \label{fig:example2}
\end{figure}

\begin{figure}[!hbt]
    \centering
    \includegraphics[width=1\linewidth]{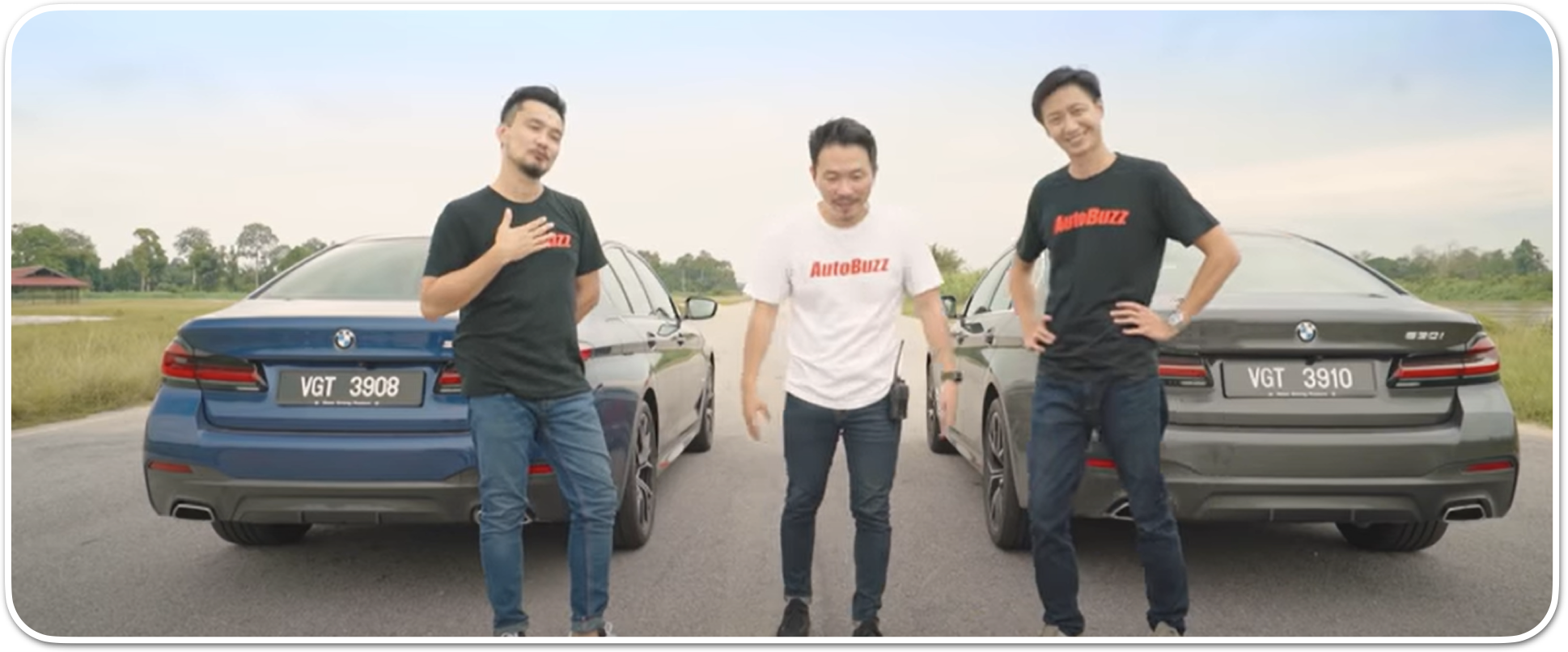}
    \caption{Example for prompt 2 featuring a BMW blue car.}
    \label{fig:example3}
\end{figure}

\begin{figure}[!hbt]
    \centering
    \includegraphics[width=1\linewidth]{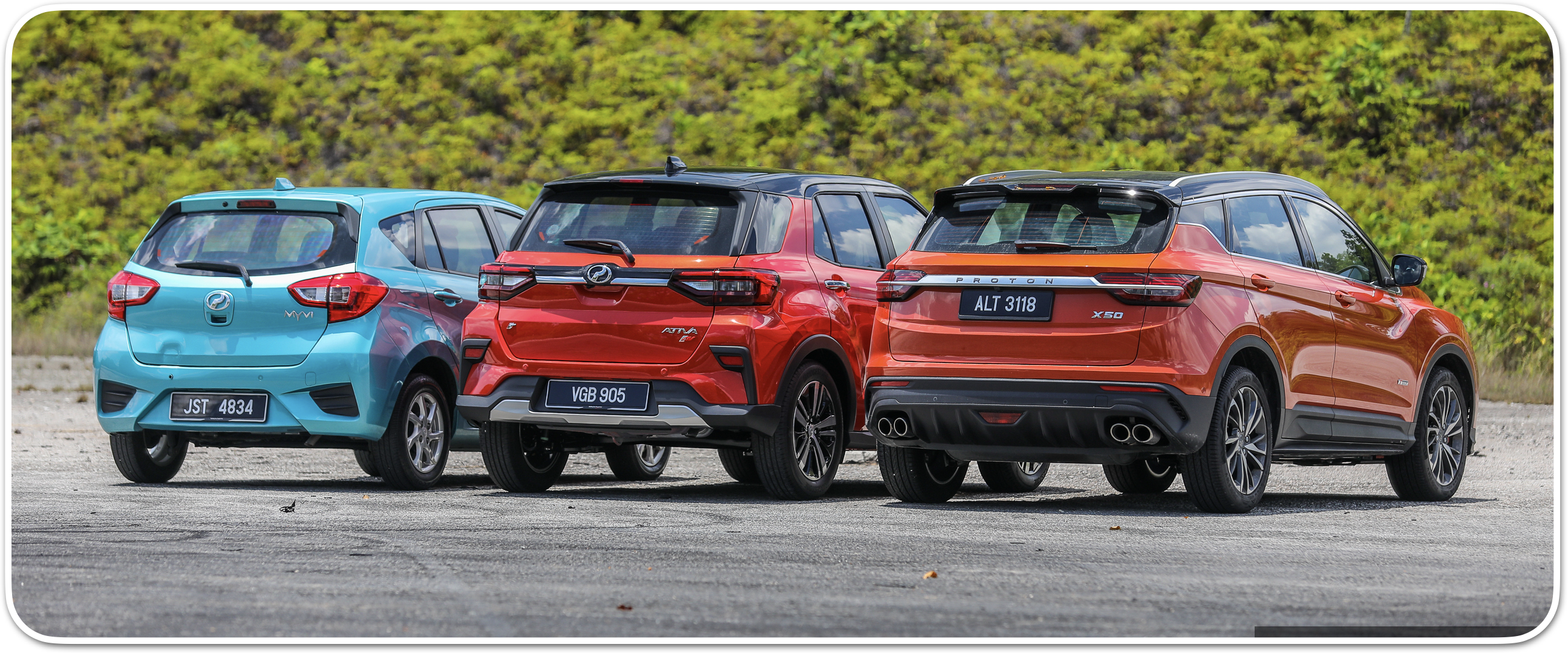}
    \caption{Example for prompt 1 and prompt 3 featuring a red car and a PERODUA car, respectively.}
    \label{fig:example4}
\end{figure}

Secondly, we evaluated both VehicleGPT and VehiclePaliGemma in several additional scenarios using other prompts, as follows:

\begin{itemize}
    \item Prompt1: ``\texttt{Extract all characters from plates of red cars}''.\\ 
    This prompt targets cars in Figure~\ref{fig:example2} and Figure~\ref{fig:example4}. 
    \item Prompt2: ``\texttt{Extract all characters from plates of BMW blue\\cars}''. 
    This prompt targets cars in Figure~\ref{fig:example3}.
    \item Prompt3: ``\texttt{Extract all characters from plates of PERODUA\\cars}''. 
    This prompt targets cars in Figure~\ref{fig:example4}.
\end{itemize}

Both VehicleGPT and PaliGemmaGPT show superior performance and produces accurate outcomes in these scenarios. This experiment underscores their ability to link the description provided in the prompts with the objects' attributes in the image to identify the specific cars' model and/or color, localize the cars and then the plates, and extract the characters from the plates.

\newpage
The strength of both VehicleGPT and and VehiclePaliGemma lies in its multitasking ability, allowing it to perform several functions simultaneously, including car localization, license plate localization, the car's model recognition, color recognition, and plate recognition. All of these functions can be driven by a prompt provided to the model along with an image. By combining multiple tasks into a single processing pipeline, organizations can save on computational costs and reduce the need for separate models for each task.

The challenging problems that VehicleGPT and VehiclePaliGemma were able to address are:
\begin{enumerate}
    \item Recognizing all cars' plates in the images, which had several cars and/or plates.
    \item Identifying multiple license plates that appeared at various angles and orientations due to the different positions and movements of the cars in real-life image captures.
    \item Being robust against the presence of various objects and textures in the background.
\end{enumerate}

\section{Conclusion and Future Work}
\label{conclusion}
This paper demonstrated the challenging problem of recognizing unclear, distorted license plates with close characters. Various VLMs have been explored to evaluate their OCR capability. We compared these VLMs with other baseline methods, utilizing a dataset of 258 pictures of Malaysian car plates. The experimental results showed that the OCR capabilities of VLMs outperformed other OCR baseline methods in terms of plate-level recognition accuracy. It was found that 226 plate images out of 258 images were recognized correctly with a plate accuracy of 87.6\% using VehiclePaliGemma, which showed superior performance compared to others. Additionally, the VehiclePaliGemma was able to correctly recognize 1710 characters out of 1751 characters with a character-level accuracy of 97.66\%.  In summary, While both VehiclePaliGemma and VehicleGPT offer excellent recognition performance, VehiclePaliGemma distinguishes itself with superior speed, affordability, and efficiency, which opens door to integrate it on edge devices for real-life scenarios. 
Moreover, we explored the multitasking capability of both ``VehicleGPT'' and ``VehiclePaliGemma'' to recognize plates in challenging conditions given an image that has multiple cars with various models and colors, as well as plates in several positions and orientations in cluttered backgrounds.

This work focused on recognizing close characters in unclear Malaysian license plates. In future work, we plan to extend the proposed solution to recognize more complex plates in other countries. Furthermore, we plan to modify the prompt to address specific instances of plates that require individual handling.  

To enhance the proposed solution and ensure no car or plate is missed, future work could involve fine-tuning PaliGemma for car and plate detection tasks. Additionally, the current solution involves multiple steps, including detecting cars and plates, recognizing the color and model of cars, and then identifying the cars. Even though all these steps are completed in under one second, further improvement could be achieved by fine-tuning PaliGemma to directly recognize plates from images containing multiple cars. However, this would require annotating a large dataset to achieve the desired performance. Such tuning should ensure that VLMs are fine tuned on diverse and representative datasets and should consider ethical implications to prevent bias and maintain privacy and security in processing such potentially sensitive information.

%-----------REFERENCES--------------
\bibliographystyle{naturemag}
\bibliography{sample}

\end{document}